\documentclass[sigconf,authorversion,nonacm]{acmart}
\AtBeginDocument{%
  \providecommand\BibTeX{{%
    \normalfont B\kern-0.5em{\scshape i\kern-0.25em b}\kern-0.8em\TeX}}}

\usepackage[inline]{enumitem}

\copyrightyear{2023}
\acmYear{2023}
\setcopyright{rightsretained}
\acmConference[GECCO '23 Companion]{Genetic and Evolutionary Computation
Conference Companion}{July 15--19, 2023}{Lisbon, Portugal}
\acmBooktitle{Genetic and Evolutionary Computation Conference Companion
(GECCO '23 Companion), July 15--19, 2023, Lisbon,
Portugal}\acmDOI{10.1145/3583133.3590703}
\acmISBN{979-8-4007-0120-7/23/07}

\begin{document}

\title{ Eco-evolutionary Dynamics of Non-episodic Neuroevolution in
Large Multi-agent Environments}

\author{Gautier Hamon}
\email{gautier.hamon@inria.fr}
\affiliation{%
  \institution{Flowers Team, Inria / Ensta ParisTech}
  \city{Bordeaux}
  \country{France}
}

\author{Eleni Nisioti}
\email{eleni.nisioti@inria.fr}
\affiliation{%
  \institution{Flowers Team, Inria / Ensta ParisTech}
  \city{Bordeaux}
  \country{France}
}

\author{Clément Moulin-Frier}
\email{clement.moulin-frier@inria.fr}
\affiliation{%
  \institution{Flowers Team, Inria / Ensta ParisTech}
  \city{Bordeaux}
  \country{France}
}

\renewcommand{\shortauthors}{}




\begin{abstract}
Neuroevolution (NE) has recently proven a competitive alternative to learning by gradient descent in reinforcement learning tasks. However, the majority of NE methods and associated simulation environments differ crucially from biological evolution: the environment is reset to initial conditions at the end of each generation, whereas natural environments are continuously modified by their inhabitants;  agents reproduce based on their ability to maximize rewards within a population, while biological organisms reproduce and die based on internal physiological variables that depend on their resource consumption; simulation environments are primarily single-agent while the biological world is inherently multi-agent and evolves alongside the population. In this work we present a method for continuously evolving adaptive agents without any environment or population reset. The environment is a large grid world with complex spatiotemporal resource generation, containing many agents that are each controlled by an evolvable recurrent neural network and locally reproduce based on their internal physiology. The entire system is implemented in JAX, allowing very fast simulation on a GPU. We show that NE can operate in an ecologically-valid non-episodic multi-agent setting, finding sustainable collective foraging strategies in the presence of a complex interplay between ecological and evolutionary dynamics. 
\end{abstract}



\maketitle

\begin{figure*}
\begin{minipage}{0.95\textwidth}
    \begin{center}
        \includegraphics[width=\textwidth]{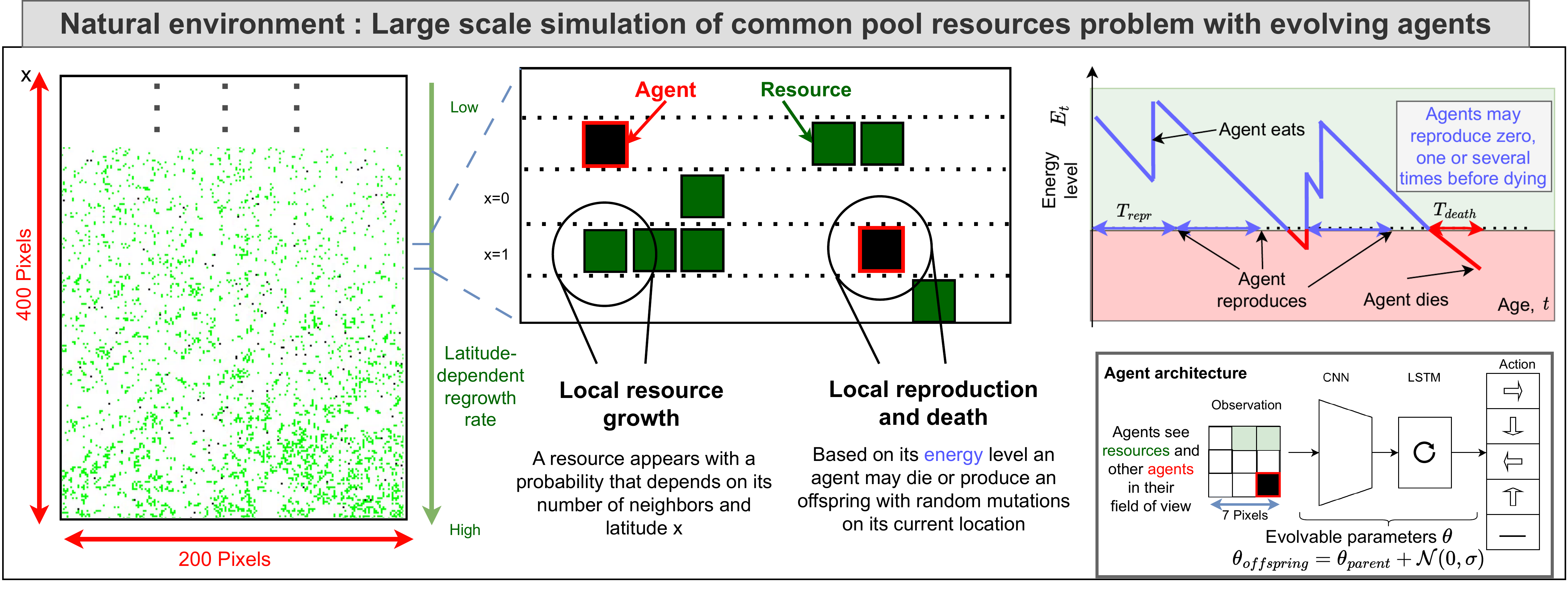}
        \caption{Our simulation environment (Left) is an extension of the Common Pool Resource (CPR) environment \citep{perolatMultiagentReinforcementLearning2017,leiboMultiagentReinforcementLearning2017a} : a two-dimensional grid-world where some cells contain resources (in green) that the agents (in black) can collect. Resources grow depending on the presence of other resources around them (local growth, Middle) with an additional very sparse spontaneous growth, which means that over-consumption may lead to their local depletion. We introduce a latitudinal model of resource regrowth similar to \cite{nisiotiPlasticityEvolvabilityEnvironmental2022} with higher resource regrowth rate in lower latitudes (bottom of the map) than in higher ones. We consider a very large environment of $200\times 400$ pixel that can contain thousands of agents (in black). We prevent any environment and population reset during a whole simulation of $1.000.000$ time steps, enabling continual eco-evolutionary dynamics to take place. Each agent may reproduce or die according to a physiological model modulating its energy level as a function of life time and resource consumption (Top-Right). Agents reproduce according to a minimal criterion \cite{brant2017MinimalCriterionCoevolution} of maintaining energy level above a certain threshold for a certain period of time. The population size therefore varies during the simulation according to the current amount of available resources and the current ability of agents to collect them. Each agent is controlled by a recurrent artificial neural network which takes as input limited local observations and outputs the navigation action (Bottom-Right). Evolution occurs through the mutation of a parent's network weights when it produces an offspring. We refer to appendix B for more details on the environment design.}
        \label{fig:envs}
    \end{center}
    \end{minipage}
\end{figure*}

\section{Introduction}

The main objective of this paper is to propose a method for studying large-scale eco-evolutionary dynamics in agent-based simulations with a reasonable level of biological and ecological plausibility. For this aim, we implement a system with the following properties (see Fig.~\ref{fig:envs} for illustration).

\paragraph{Non-episodic simulation environment with complex intrinsic dynamics}. We model our environment after common-pool resource (CPR) appropriation problems, where a group of agents competes for finite resources. We extend an existing environment of CPR appropriation~\citep{perolatMultiagentReinforcementLearning2017} with the presence of multiple niches, where resources regrow proportionally to the density of nearby resources at different rates in different regions of the environment (Fig~\ref{fig:envs}). We prevent any environment or population reset during a whole simulation run, enabling coupled environmental and population dynamics leading to complex eco-evolutionary feedback effects.

\paragraph{Continuous neuroevolution in a large, size-varying agent population} The environment contains thousands of agents, each controlled by a neural network whose weights are optimized using neuroevolution \citep{stanley2019DesigningNeuralNetworks}. Each network contains a memory component (LSTM), which enables adaptation within the agent's lifetime in the absence of weight updates. Thus the evolutionary process can be viewed as an outer loop that optimizes the ability of agents to adapt to different environmental conditions.

\paragraph{Physiology-driven death and reproduction} There is no notion of rewards, agents are instead equipped with a physiological system modulating their energy level according to the resources they consume, in a non-linear way. At the evolutionary scale, agents reproduce as long as they are able to maintain their energy level within a reasonable range and die if this level goes below a minimum threshold. This is departure from the notion of fitness-based selection and more in line with a minimal criterion selection \citep{brant2017MinimalCriterionCoevolution}. Note that the population size can vary with time. 

\paragraph{Natural vs. Lab environments: Evaluation methodology}

As we are interested in the system's ability to emerge interesting behaviors that hint to open-ended dynamics, evaluating it on pre-defined set of tasks would defeat our purpose. For this reason we have structured our simulation methodology as follows: we let the population of agents evolve for a long time in a single environment and study its behavior at a large global scale and at a smaller local scale. At the large scale, we study the dynamics of the system in what we call the "natural environment", i.e. the full simulation run,  by monitoring population-wide and terrain-wide metrics. At the small scale, we first focus on local, interesting patterns of behaviors observed in the natural environment, such as individual agents that move in a consistent way or collective immigration and foraging patterns. We then form specific hypotheses about the potential drives of these behaviors and evaluate selected agents in specific "lab environments" that enable testing these hypotheses. These environments differ from the one used for evolving behaviors: they are much smaller and exhibit vastly different population and resource dynamics (we illustrate  examples of such environments in Figure \ref{fig:all_results}.E).

From the perspective of neuroevolution, our empirical study aims at answering the following questions: a) \textit{can we realistically apply neuroevolution in multi-agent environments with thousands of agents?} b) \textit{does a selection mechanism that allows agents to reproduce locally, without requiring generational resets, based on a minimal criterion suffice?} c) \textit{does evolving networks in a multi-agent setting lead to the emergence of adaptation mechanisms?} From the perspective of multi-agent cooperation, our study targets the questions: a) \textit{can we simulate systems with complex eco-evo dynamics where populations solving a CPR problem exhibit realistic behaviors?} b) \textit{does evolving under a minimal criterion enable sustainability}? In the next section we answer these questions in the affirmative. 

Leveraging the GPU parallelization allowed by the JAX programming framework \citep{jax2018github}, we run large-scale continual simulations in grid-world environments with approximately $100K$ cells and thousands of agents (notably a simulation of $1M$ time steps with such a population requires about only 20 minutes). 

More detailed background positioning our contributions with respect to the existing literature, as well the computational details of our proposed method, are provided in the Appendix.

\section{Results}\label{sec:results}

 We will now study the evolution of a population in our proposed system and probe certain quantities during evolution. Note that this system required some tuning of the hyperparameters in order to find a stable environment, as exponential growth of both food and population can easily lead to collapse (and even did after several generations in 3 out of the 5 seeds launched). We will make a detailed analysis of one seed and refer to Appendix D.2 for an analysis of another one with a different eco-evolutionary path. We provide videos that show the real-time behavior of our system in a companion website (\url{https://sites.google.com/view/non-episodic-neuroevolution-in/}) as well as a repository containing code for reproducing our experiments (\url{https://github.com/flowersteam/EcoEvoJax}). 

 Details on the environment and  hyperparameters characterizing the natural  environment can be found in Appendix B  and an explanation of how the metrics have been implemented and how statistical significance was tested for in Appendix C. 
 
\subsection{Eco-evolutionary dynamics}

 In this simulation, the evolution of the population is deeply interconnected with the evolution of resources. In Fig \ref{fig:all_results}.A.2, we observe that at a small scale the population size (blue) and resources (green) present in the environment follow a predator-prey Lokta-Volterra dynamic \cite{volterraFluctuations1926}. Those oscillation are interesting from an evolutionary perspective: easier phases with higher resources availability, in which the population grows and where diversity can emerge, are followed by high competition phases due to an increase in population and decrease in resources.
 


\subsubsection{Coexistence of agents with different movement dynamics}

\begin{figure}
    \begin{minipage}{0.45\textwidth}
    \begin{center}
    \includegraphics[width=\textwidth]{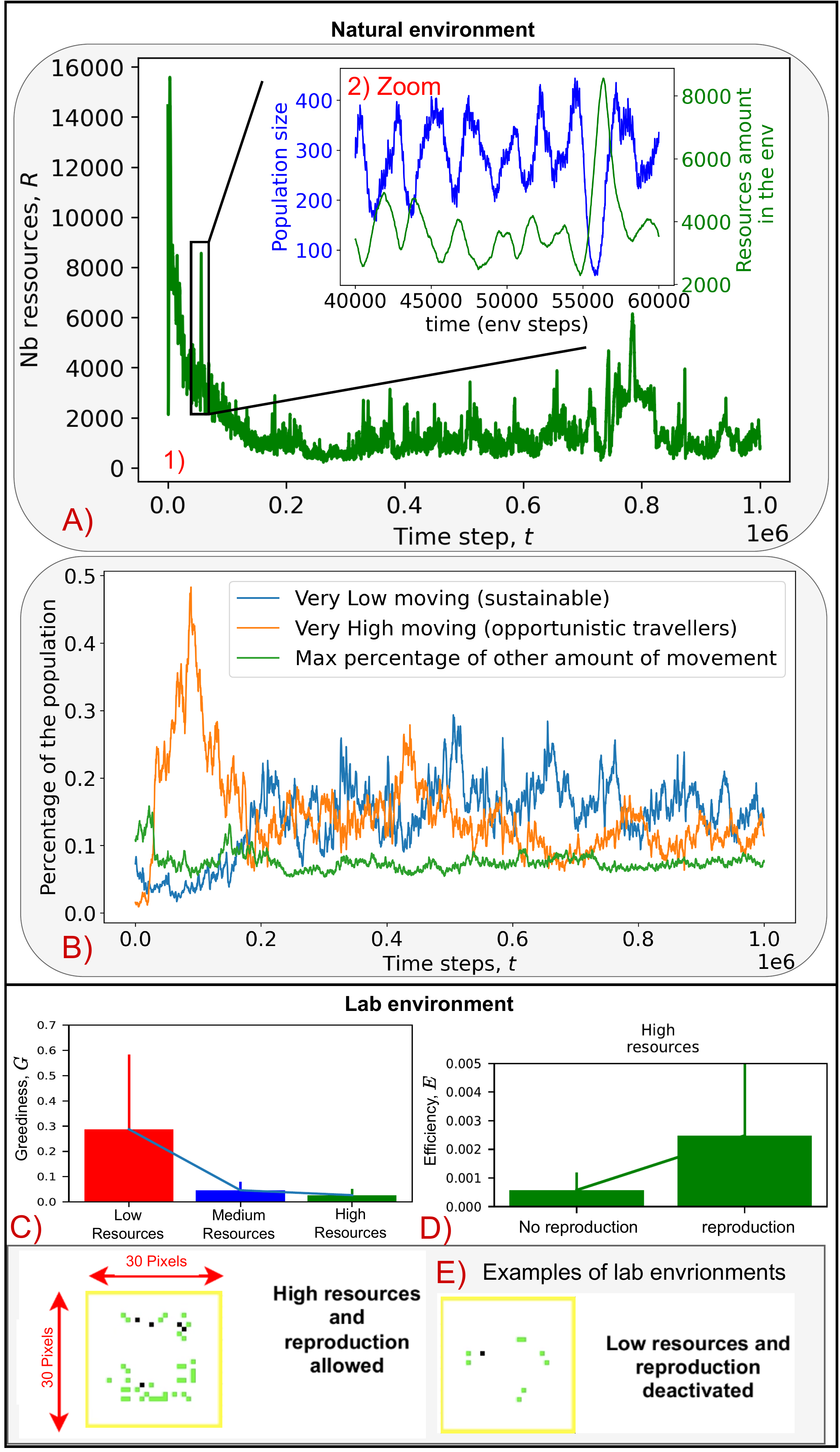}

    \end{center}
    \end{minipage}
    \caption{A) 1) Amount of resources in the environment over time and 2) zoom  on a smaller timescale showing the interplay with population size (blue) B) Percentage of individuals with different amount of movement over time C) Greediness of a sustainable forager agent across evaluation environments that differ in the amount of resources. D) Average efficiency across the population in high resources task with reproduction activated and deactivated. Activating reproduction leads to increased resource
consumption. 
    }
    \label{fig:all_results}
\end{figure}

At the beginning of evolution  (steps $0-200K$, starting with random agents), the environment has abundant resources which leads to to high-moving behaviors as an easy first strategy in this high-resource environment (Fig \ref{fig:all_results}.A,B). Then, when the amount of resource decreases, we observe an increase in the number of low-moving individuals (Fig \ref{fig:all_results}.A,B) exploiting local resource spots (from step $200K$). From this point, those two extreme strategies coexist in the agent population (Fig \ref{fig:all_results}.B). This differs from previous related work in a similar environment \cite{AubertKatoHungerGames2015}, relying on a simpler agent architecture and a fitness-based reproduction condition, where only one strategy ended up populating the whole environment. Those extreme behaviors correspond to two distinct types of agents: high movement individuals are agents that have an "opportunistic traveler" strategy as they travel mostly in straight line but opportunistically exploit resources spots locally (especially from isolated resources from the sparse spontaneous growth) as soon as they see them. On the other hand, the low movement individuals exploit the spreading of resources by staying at the same interesting place (with resources around) and waiting for resources to spread. We qualify this waiting of resources as a sustainable strategy as agents do not consume resources greedily but rather keep these resources as a reliable source of respawn for more long term survival (for themselves but also for their offspring that will inherit this place). We refer to video 1.a of the 
\href{https://sites.google.com/view/non-episodic-neuroevolution-in/}{\underline{companion website}} for a visualization of these behaviors and to the next subsection for a more detailed and controlled analysis of the behavior (and diversity) of agents.


\subsection{Evaluation in lab environments}
How do the agents adapt their foraging behavior at an evolutionary and intra-life timescale to maximize their reproduction rate? In the natural environment we saw that both population size and the spontaneous regrowth of resources may contribute to avoiding resource depletion. At an evolutionary scale the population may adapt by regulating its size and updating its weights. But is it possible that the agents learned to adapt to different conditions they encounter in their lifetime in order to forage both efficiently and sustainably? This is the question the following simulations in the lab environments aim to address.

\subsubsection{Does the density of resources affect agents' greediness?}

\paragraph{Set-up} There are three lab environments, with a single agent that cannot reproduce and resource regeneration deactivated, that differ in the amount of initial resources (see on Figure \ref{fig:all_results}.E for an illustration of the low and high resources environments). In each of these lab environment, we measure the amount of greediness $G$, by dividing the simulation into non-overlapping fixed windows of 20 timesteps and checking in which of these windows the agent has at least one resource in its field of view (let's denote this number with $T_r$) and the number of these windows during which the agent consumed at least one resource (let's denote this number with $C_r$), so that $G=C_r/T_r$  . We compute this measure on randomly sampled evolved agents from the end of the natural environment simulation.  To quantify the effect of the density of resources we perform statistical tests comparing the greediness of each agent in the three tasks. More information can be found in Appendix C.

Our analysis showed that agents exhibit different qualitative behaviors that can be grouped in two types: a) agents for which no statistically significant differences appear between tasks. These agents correspond to the \textit{opportunistic travelers} that we encountered in the natural environment and do not exhibit resource-dependent adaptation b) agents for which there are statistically significant differences between the low-resources and high-resources environment, with greediness in low-resource environments being higher. Overall, 9 out of the 50 agents exhibited this behavior (we illustrate greediness across tasks for one of these agents in Figure \ref{fig:all_results}.C), which we refer to as \textit{sustainable foragers}. These agents have learned to not over-consume resources when these are abundant, but stay close to them to consume them later and take advantage of the higher spread rate. On the other hand, low resources environment means slower spread (and may mean lattitude with lower regrow) which might exlain why even those sustainable agent prefer to take the resource and leave. 



\subsubsection{Does peer-pressure lead to greediness?}

\paragraph{Set-up} We use the high resources task but now allow agents to reproduce. This means that, after $T_{repr}=20$ timesteps, new agents will appear, leading to competition for resources. Our hypothesis is that this will make agents more greedy. To test this, we measure efficiency $E$ as the average amount of resources every individual consumes during the evaluation trial and average it across 50 agents and 10 trials. We then compare the difference in performance between the previous set-up (no-reproduction) with the current one, where we average across 50 agents and 10 trials to observe whether there is a population-wide effect.

As Figure \ref{fig:all_results}.D illustrates, we observed a large change in the foraging efficiency of the agents when reproduction was on. Efficiency increased by a statistically significant amount, which indicates the sustainable foragers increased their greediness under peer pressure. However, we observed that, after an initial increase in resource consumption at the appearance of new agents, the group slows down again and its members tend to disperse and stay close to resources without consuming them (see \href{https://sites.google.com/view/non-episodic-neuroevolution-in/}{\underline{companion website}}.B.1  for an illustration of this behavior).

\section{Discussion}
Our empirical study demonstrates that neuroevolution can operate in large multi-agent environments, lead to efficient behaviors even in the absence of episodic survival-of-the-fittest and help evolve agents that exhibits adaptation within their lifetime without requiring weight updates. Specifically in regards to the latter, we identified agents that change their policy depending on resource density and presence of other agents. From an ecological perspective, our computational study proves that agents selected based on a minimal criterion learn sustainable behaviors and that the population exhibits dynamics that resemble those of natural populations, such as population size oscillations. We observed many interesting emerging examples of collective and individual adaptation, including: 
\begin{enumerate*}[label=\roman*)]
    \item Population size exhibits bursts and crashes that are correlated with the density of resources,
    \item The system goes through phases related to the sustainability of the agents' foraging behavior: resources and population size initially grow until over-population leads to near-extinction of resources which creates a drive for agents to forage sustainably,
    \item The sustainable population exhibits diversity in individual behaviors: some agents specialize in long-distance travel, opportunistically consuming resources they find on their way, while others forage locally, staying close to resources to take advantage of the spread of resources and consuming sporadically to avoid death,
    \item Agents' influence each others behavior: agents that forage sustainably when alone, temporarily  increase their consumption when others enter their field of view and then revert back to consuming less.
\end{enumerate*}

Interestingly, points i) and ii) above could not be observed in a standard episodic training paradigm, where environment and population resets would prevent any eco-evolutionary feedback. 
In this respect, we are considering future experiments studying whether continual local reproduction, where offsprings are produced next to their parent, did enable some sort of kin selection --e.g. in the form of reducing parent's greediness as a way to favor the survival of their offsprings. Other future work could also focus on studying to what extent the memory component of the agent's cognitive architecture contributes to intra-life adaptation. 

In the past ecologists have hinted to the limitations of an anthropocentric view on intelligence \citep{dewaal2016AreWeSmart}: if we search for intelligence by looking at performance metrics only in tasks that we excel at,  then we will inevitably miss a big part of the natural kingdom. Our study hints to a similar conclusion for artificial agents: evolving agents in natural environments with complex spatiotemporal dynamics in the absence of rewards and examining their behavior in toy lab environments may bring us closer to our quest for open-end behavior in artificial systems.

\bibliographystyle{ACM-Reference-Format}
\bibliography{sample-base}

\pagebreak

\appendix

\section{Background}\label{sec:background}


There are striking differences in how adaptation operates in biological versus artificial systems. In Artificial Intelligence (AI), the most common approach is \emph{performance-driven}. 
The main assumption is that intelligence must be implemented in a structured cognitive architecture (integrating e.g. control, learning and memory mechanisms) which is optimized (using machine learning methods) through pre-defined objective functions \citep{chollet2019MeasureIntelligence,marcus2019RebootingAIBuilding,silver2021RewardEnough}. The proposed methods are evaluated in benchmarks designed to capture various aspects of intelligence. For example, \citet{chollet2019MeasureIntelligence} defines intelligence as \textit{a measure of its skill-acquisition efficiency over a scope of tasks, with respect to priors, experience, and generalization difficulty} and proposes a benchmark to evaluate it inspired by psychometric intelligence tests and called the \textit{The Abstraction and Reasoning Corpus (ARC)}. The rise of deep neural networks as powerful function approximators has strongly revived this approach by allowing key advances in e.g. representation learning and reinforcement learning in high-dimensional spaces~\citep{mnih2015HumanlevelControlDeep}. 

In contrast, biological adaptation seems to be better characterized by the notion of open-endedness (the continual generation of increasingly diverse organisms) than by the notion of performance. While the popular concept of the \emph{survival of the fittest} suggests that biological evolution is driven by performance, this concept has actually little grounding in evolutionary theory~\citep{stanley2015WhyGreatnessCannot}. An important paradigm shift is taking increasing importance in evolutionary biology, recognizing the crucial role of eco-evolutionary feedbacks as a main driver of evolution. The \emph{extended evolutionary synthesis}~\citep{pigliucci2010evolution,laland2015ExtendedEvolutionarySynthesis} considers that, in the standard evolution theory as well as its modern synthesis, \textit{too much causal significance is afforded to genes and selection, and not enough to the developmental processes that create novel variants, contribute to heredity, generate adaptive fit, and thereby direct the course of evolution}. It recognizes important feedback effects in terms of \emph{constructive development}, i.e. the ability of an organism to shape its own developmental trajectory by constantly responding to, and altering, internal and external states, as well as of \emph{reciprocal causation}, i.e. that developing organisms are not solely products but, by modifying their niche and therefore its associated fitness landscape, are also causes of evolution.

Following a similar paradigm shift, a recent trend in AI is increasingly recognizing the importance of reciprocal influence between adaptation and environmental dynamics~\citep{clune2020AIGAsAIgeneratingAlgorithms,stanley2015WhyGreatnessCannot,leibo2019AutocurriculaEmergenceInnovation}. This approach, that we can qualify as \emph{complexity-driven} (by opposition to the \emph{performance-driven} approach mentioned above), considers intelligence as the emergent product of adaptive systems interacting with complex environmental dynamics. There are two main propositions here. Some contributions study how competition and cooperation pressures in populations of co-adapting agents can result in a behavioral arms race where each agent has to continuously improve its skills against those of other agents, an approach called \emph{multi-agent autocurriculum} \citep{leibo2019AutocurriculaEmergenceInnovation}. Other contributions study how learning algorithms themselves can be meta-learned for operating in a diversity of environments. \citet{clune2020AIGAsAIgeneratingAlgorithms} calls this approach \emph{AI-Generating Algorithms (AI-GA)}, with three main pillars: \textit{(1) meta-learning architectures, (2) meta-learning the learning algorithms themselves, and (3) generating effective learning environments}. In both propositions (autocurriculum and AI-GA), it is the complexity of the environment (either through the presence of other co-adapting agents or through its intrinsic diversity) that drives the ability to continuously acquire new skills and generalize them in novel environments. In other words, if the performance-driven approach attempts to reverse-engineer the brain (or at least its main functions), the complexity-driven approach instead attempts to reverse-engineer the environmental conditions that lead to intelligence.

Both performance-driven and complexity-driven approaches, however, still differ crucially from how adaptation occurs in the natural world. Even complexity-driven approaches are almost exclusively evaluated in terms of their convergence toward efficient policies on evaluation tasks (e.g. their performance in board games against humans~\citep{silverMasteringGameGo2017}, or in generalization to novel test environments~\citep{wang2019PairedOpenEndedTrailblazer}). This is especially due to the fact that they mostly rely on the reinforcement learning (RL) framework.  A central assumption in RL is that the objective of an artificial agent must be to learn a task (or a set of tasks), that these tasks should be defined as reward functions, and that those rewards are provided by the environment. From a biological perspective, however, the environment does not contain any reward whatsoever. Rewards instead, if at all, result from the agent's own physiology and self-regulation and have emerged from evolution as a way to guide learning and exploration (see e.g. the metaphor of evolved stick/carrot mechanisms in~\citep{sterling2012AllostasisModelPredictive}). Second, the standard training paradigm in RL is episodic, i.e. the environment is regularly reset to its initial conditions. While this procedure has the benefit to facilitate training from a machine learning perspective~\citep{pardo2018TimeLimitsReinforcement}, it strongly differs from natural settings where environments are persistent, i.e. where the behavior of agents affects the environment in which the next generations will further evolve and learn (see however~\citep{co-reyes2020EcologicalReinforcementLearning} for a recent attempt at studying ecologically valid non-episodic RL settings). Episodic training in RL prevents the study of both niche construction and eco-evolutionary feedback effects, which require that populations alter their environment and that those changes in the environment influence the subsequent evolution of the population~\citep{post2009EcoevolutionaryFeedbacksCommunity}. Contributions in automatic curriculum learning~\citep{portelas2020AutomaticCurriculumLearning} studies such feedback effects but focuses on how to adaptively sample novel environments of increasing complexity, using episodic training, with the explicit objective to improve an agent's learning performance.

\subsection{Neuroevolution}
Neuroevolution draws inspiration from natural evolution to create agents that learn to adapt through an evolutionary process rather than gradient-based optimization \citep{stanley2019DesigningNeuralNetworks}. In a surprise to many, this simple process of selection and random mutations has recently performed competitively with the state-of-the-art art in RL for playing Atari games \citep{salimans2017EvolutionStrategiesScalable,such2018DeepNeuroevolutionGenetic}, and proven powerful in applications such as architecture search, where the non-differentiable nature of the search space prohibits gradient-based methods
\citep{liu2023SurveyEvolutionaryNeural} and meta-learning, where the evolutionary process is conceived as an outer optimization loops that controls the intra-life learning plasticity of agents \citep{risi2010EvolvingPlasticNeural}. Multi-agent environments, which are particularly promising for neuroevolution as they naturally entail the concept of a population, have been identified as a frontier for this family of methods \citep{miikkulainenMultiagentLearningNeuroevolution2012}, arguably due to their computational complexity and challenging multi-agent learning dynamics . 

Neuroevolution methods are classically performance-driven: solutions are selected based on their ability to solve a pre-determined task. Complexity-driven approaches, on the other hand, where solutions are chosen based on criteria not directly related to performance, such as novelty, have proven powerful in tasks for which the objective function is unknown to humans \citep{woolley2011DeleteriousEffectsPriori}. For a given criteria, neuroevolution methods can also differ on whether solutions survive only if they are ranked high within the population (survival of the fittest) or if their fitness is above a threshold (minimum criterion). The latter category is the least explored \citep{brant2017MinimalCriterionCoevolution}, but has the potential of preserving a larger phenotypic diversity within the population and is believed to be closer to biological evolution.

Finally, neuroevolution methods almost exclusively consider discrete, over-lapping generations, at the beginning of which solutions experience mutation and selection simultaneously and the environment is reset to its initial conditions. We refer to this paradigm as episodic, borrowing terminology from RL, where recently it has been proposed to remove environmental resets, as they may introduce the need for human supervision \citep{gupta2021ResetFreeReinforcementLearning} and are implausible from a biological perspective \citep{co-reyes2020EcologicalReinforcementLearning}. This setting, termed as non-episodic or continuous in RL, is harder to envision in evolution under survival-of-the-fittest, where dividing time into non-overlapping generations ensures that agents compete based on the same time budget. 

\subsection{Common-pool resource appropriation}

CPR tasks abide in natural and human ecosystems: fisheries, grazing pastures and irrigation systems are examples of multi-agent systems where self-interested agents need to reach a sustainable resource appropriation strategy that does not exploit the finite resources. They belong to a class of game-theoretic tasks termed as social dilemmas, which exhibit a tension between individual and collective motives: the optimal collective strategy is to forage sustainably but self-interested agents will cooperate only if others cooperate as well; otherwise they will consume resources until they deplete them, a situation called Tragedy of the Commons \citep{hardin1968TragedyCommons}. Ecological properties of these complex systems, such as the spatiotemporal variability of resources and organisms are believed to play a big part in shaping solutions to CPR problems  \citep{janssen2010IntroducingEcologicalDynamics}. From an ecological perspective, such settings give rise to scramble competition, where organisms of the same species appropriate resources at a rate contingent on their foraging ability, often leading to population bursts and crashes  \citep{parker2000ScrambleBehaviourEcology}. 

With recent advances in RL, computational studies of social dilemmas have managed to operate in simulation environments resembling the ones used in human lab studies, where agents can navigate a grid-world consuming resources \citep{perolatMultiagentReinforcementLearning2017,leiboMultiagentReinforcementLearning2017a}. RL agents embody the self-interested trial-and-error learning paradigm and have confirmed our intuition that, when acting in a group, they cannot avoid a Tragedy of the Commons unless they employ some auxiliary mechanism for guarding against exploiters, such as learning to incur punishment \citep{perolatMultiagentReinforcementLearning2017} and reputation mechanisms \citep{anastassacosCooperationReputationDynamics2021a}. These studies, however, remain far from approaching the complexity of real ecosystems, which may comprise thousands of organisms that do not necessarily follow the reward-maximization paradigm.   

\section{Details of the simulation}\label{appendix:details_simu}

\subsection{Environment}

Our simulation environment is an extension of the CPR environment \citep{perolatMultiagentReinforcementLearning2017,leiboMultiagentReinforcementLearning2017a} that the AI community has been using to study the emergence of cooperation in groups of self-interested agents: a two-dimensional grid-world where some cells contain resources (in green) that the agents (in black) can collect. Resources grow depending of the presence of other resources around them, which means that there is a positive feedback loop, with reduction in resources leading to further reductions. In addition to resources, the environment may contain walls (in blue) that kill agents trying to traverse them (see Figure \ref{fig:envs} for an illustration of our environment).

At each time step $t$ of the simulation a resource may grow in a pixel in a cell of the environment with location $(x,y)$ based on the following three processes:

\begin{itemize}
    \item a neighborhood-dependent probability $p_{I}(x,y)$ determines the probability of regrowth in a cell based on the number of resources in its neighborhood, $I$
    \item a niche-dependent scaling factor $c(x)$ is used to scale $p_{I}$. We employ a latitudinal niching model used in previous studies \citep{nisiotiPlasticityEvolvabilityEnvironmental2022,groveEvolutionDispersalClimatic2014a}: the world is divided into $N$ niches, each one having the form of a horizontal stripe of pixels so that a cell's location depends only on its vertical position $x$. We refer to $c(x)$ as the climate value of niche $x$.
    \item independently of its neighbors and niche, a resource grows with a  constant low probability $c$. This is what we refer to as (sparse) spontaneous growth.
\end{itemize}

By modeling resource generation in this way we ensure that the resource distribution follows the CPR model, that it exhibits additional spatio-temporal variability due to the presence of niches and that resources do not disappear too easily, which can be problematic in reset-free environments. Thus, the combined regrowth rate for a resource $r$ is:

\begin{align}
    p(x,y) = p_I(x,y) \cdot c(x) + c
\end{align}

A niche's climate value is determined by equation: $ c(x) = (\alpha^x +1)/(\alpha + 1)$, which returns values from 0 to 1 and allows us to control the relationship between niche location and climate to be from linear to exponential.

\subsection{The agents}
At each time step there is a variable number of agents $K_t$ in the environment,  each one characterized by its sensorimotor ability, cognitive capacity and physiology.

\paragraph{Sensorimotor ability} An agent observes pixel values at each time step within its visual range (a square of size [$w_o$, $w_o]$ centered around the agent, as illustrated in the bottom right part of Figure \ref{fig:envs}). The pixel values contain information about the resources, other agents (including their number) and walls. At each time step an agent can choose to stay inactive or execute an action to navigate up, down, right or left.

\paragraph{Cognitive capacity} An agent is equipped with an artificial neural network that outputs the action to undertake based on the current observation and whose weights are initialized randomly once at the start of the simulation. Its architecture (illustrated in the bottom right part of Figure \ref{fig:envs}) is minimal: a convolutional neural network, an LSTM cell that equips the agents with memory by enabling policies conditioned on a trajectory of observatories and a linear layer that transforms hidden states to actions.

\paragraph{Physiology} An agent is equipped with a simple physiological model modulating its level of energy: the agent is born with an initial energy value $E_0$ which, at every time step, experiences a linear decrease, and, if the agent consumes a resource, is increased by one (see the top right part of Figure \ref{fig:envs} for an illustrative example of how the energy level may change within the lifetime of a hypothetical agent). The energy is  also clipped to a max value $E_{max}$.

\subsection{Non-episodic neuroevolution}
In neuroevolution (NE) a population of neural networks adapts its weights through random mutations and a selection mechanism that promotes well-performing policies. Under a classical NE paradigm training time is divided into generations, at the end of which agents reproduce to form the next generation \citep{stanley2019DesigningNeuralNetworks,salimans2017EvolutionStrategiesScalable}. 

Our proposed system deviates from this paradigm in two respects:

\begin{itemize}
    \item agents do not reproduce according to their fitness but according to a minimal criterion~\cite{brant2017MinimalCriterionCoevolution,brant2020DiversityPreservationMinimal} on their energy level;
    \item evolution is non-episodic: upon satisfying certain criteria an agent reproduces locally (the off-spring appears on the same cell as its parent), so that agents are added in an online fashion to the population, removing the need for a concept of generation.
\end{itemize}

\paragraph{Reproduction} In order to reproduce an agent needs to maintain its energy level above a threshold $E_{\text{min}}$ for at least $T_{\text{repr}}$ time steps. Once this happens the agent produces an off-spring and is a candidate for reproduction again. Thus, agents may have a variable number of off-spring and do not die upon reproduction. We illustrate this relationship between energy level and reproduction in the top right part of Figure \ref{fig:envs}. Reproduction is asexual: an agent's weights are mutated by adding noise sampled from $\mathcal{N}(0,\sigma)$

\paragraph{Death} An agent dies once its energy level has been below a threshold $E_{\text{min}}$ for at least $T_{\text{death}}$ time-steps or if its age is bigger than a certain value $L_{\text{max}}$. Once this happens, the agent is removed from the population forever.

\subsection{Hyperparameters of the simulation}

Below we provide the values of all hyper-parameters of the environment:

\begin{itemize}
\item grid size : 400x200
\item Max population :1000 agents
\item Starting population : 330 agents randomly placed
\item Starting resources : 16 0000 randomly placed resources (there can only be at most 1 resources at one pixel)
\item Field of view of agents $w_o$ : 15 (15x15 square with 7 in every direction)
\item Total number of timesteps : 1e6
\item Mutation variance $\sigma$ : 0.02
\item Energy function parameters :
    \begin{itemize}
    \item Time to reproduce $T_{reproduce}$ : 140 timesteps
    \item Time to die $T_{death}$: 200 timesteps
    \item Max energy $E_{max}$ = Starting energy $E_0$ : 3 
    \item Energy death $E_{death}$ : 0
    \item Energy decay : 0.025
    \item Increase in energy when eating a resource : 1
    \item Maximum age : 650
    \end{itemize}
\item Regrowth function
    \begin{itemize}
        \item $p(x,y) = p_{I}(x,y) \cdot c(x) + c$
        \item $p_{I}(x,y) = \mathbb{1}_{I=1}* 0.002 $
        \item  $I$ corresponds to the 4 direct neighbors 
        \item $c(x) = \frac{(\alpha^x +1)}{(\alpha + 1)}$ with $\alpha$ = 200
        \item c = 0.00005
    \end{itemize}

\end{itemize}

The simulation of the environment with a large number of agents and 1e6 timesteps took 20 minutes on a single GPU thanks to JAX parallelization and speedup.

\subsection{Details of agents architecture}

The observation of the agent is fed into a 2 layer Convolutional neural network (CNN) : First CNN has a kernel size (3,3), stride 2 and number of features 4 followed by an average pooling of size (2,2) and stride 1 , the second CNN has a kernel size (3,3, stride 2 and number of features 8 followed by an average pooling of size (2,2) and stride 1. The output of the CNN is then flattened and we concatenate to this vector the previous action of the agent as well as a binary telling if the agent has eaten a resources or not. This embedding is then fed into the LSTM of hidden state size 4 and we concatenate the embedding with the output of the LSTM. The vector obtained is then fed into a dense layer of size 8 and a tanh activation function. We finally apply a last denser layer of size 5 (the number of actions) with softmax to get the action probability from which the action will be sampled. The softmax we use has a low temperature (1/50), so that the evolution can quickly learn non random policy. 

In total the agent neural network is very small with only 2445 parameters. We chose a small neural network in order to make the evolution learning easier. 

\section{Details on measures and evaluation}\label{app:eval}

\begin{figure}
    \centering
    \includegraphics[width=0.45\textwidth]{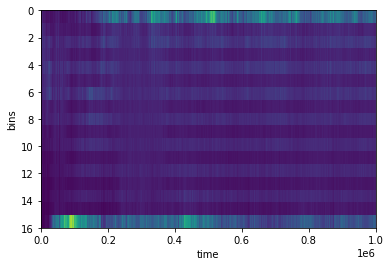}
    \caption{Heatmap of the amount of individuals in 17 bins of the distance traveled during a time window in seed 1. y axis corresponds to the bins while x axis is the timesteps of the natural env simulation. On the bin (y axis) 0 are agent that traveled a very small distance while 16 are agent that covered a long distance (nearly the maximum amount of distance you can travel during the 50 timesteps window)}
    \label{fig:heatmap_movement}
\end{figure}

The metrics used to characterize the system in the natural and lab environment are:
\begin{itemize}
    \item Amount of resources in the map : sum over the whole grid of the resource channel  
    \item Population size: Number of individual alive in the simulation
    \item Expectancy : Average of age of agent that died during a large time window of 500 timesteps.
    \item Percentage of the population with different amount of movement : For every agent alive during a time window of 50, we compute the Manhattan distance between the position at the last timestep of the window and the position at the beginning of the window. This gives the distance traveled during this time window for every agent alive as a number between 0 and 50. We then make 17 bins out of those 51 possible distance traveled. We then report in Figure \ref{fig:all_results}.B the percentage of the population in the two extreme bins as well as the max on the other ones. We provide in Figure \ref{fig:heatmap_movement} a heatmap displaying the all 17 bins during the whole natural environment simulation of seed 1. 
\end{itemize}

For statistical testing we employ the ANOVA test to detect differences across multiple conditions and Tukey's range test for pairwise comparisons. We report as statistically significant differences between pairs of methods with a p-value lower than 0.05.

\section{Additional results}

\begin{figure}

\begin{minipage}{0.45\textwidth}
    \begin{center}
    \includegraphics[width=\textwidth]{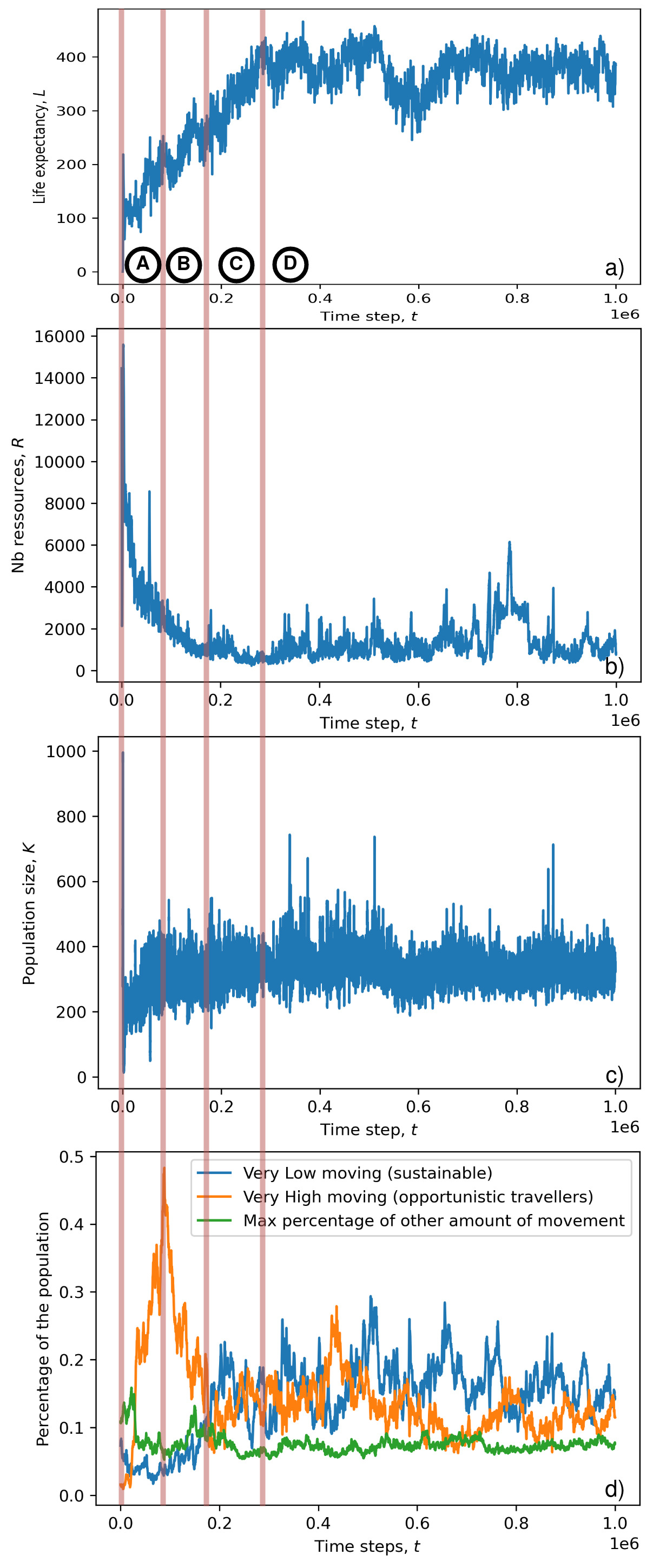}
    \end{center}
\end{minipage}
\caption{Metrics on seed 1: a) life expectancy of agents, b) total number of resources present in the environment, c) size of the population and d) percentage of individuals with different amount of movement}
\label{fig:simulation_metrics}
\end{figure}

\subsection{Large-scale trends}\label{appendix:seed1_metrics}

At the large scale several phases can be seen in the evolution of agents and the environment. In this section, we explore those phases in seed 1 (which is the seed described in the main paper in section \ref{sec:results} ) and give some new metrics of the natural environment not mentioned in the main paper such as mean life expectancy of the agents.

 \subsubsection{Population size and life expectancy rise and plateau}

 At the very beginning, in the first phase A (fig \ref{fig:simulation_metrics}.A), the environment contains plenty of resources which leads to an increase in the population. In a second phase (fig \ref{fig:simulation_metrics}.B), the population seem to start to plateau while the amount of resources is still decreasing. This decrease in resources stops in phase C.

  During phase A, B and C, the expectancy of the agents increases (fig \ref{fig:simulation_metrics}.a) suggesting that the agents are becoming better even though the environment is changing. The expectancy starts to plateau in phase D where it seems like the environment reaches a more or less stable state on some metrics. 
 
 \subsubsection{Decrease in the amount of resources: A near tragedy-of-the-commons.}
 
 The decrease in the amount of resources in the environment at the beginning (fig \ref{fig:simulation_metrics}.b), seems to indicate that the evolving population as a whole depletes the resources in a greedy way even though more resources means a higher spawn of resources. The evolutionary path therefore seems to start by evolving a population which will go towards the tragedy of the common (which is here dampened by the fact that there are sparse spontaneous growth of resources). This is confirmed by looking at the environment after some time (fig \ref{fig:evo_path}.a) where we can see that there are only few patch of resources in some corner of the map while the majority of the map is constantly depleted. This suggest that at least local tragedy of the common happens in our simulation.


\subsection{ Seed 2}\label{appendix:seed2}
\begin{figure}

\begin{minipage}{0.45\textwidth}
    \begin{center}
    \includegraphics[width=\textwidth]{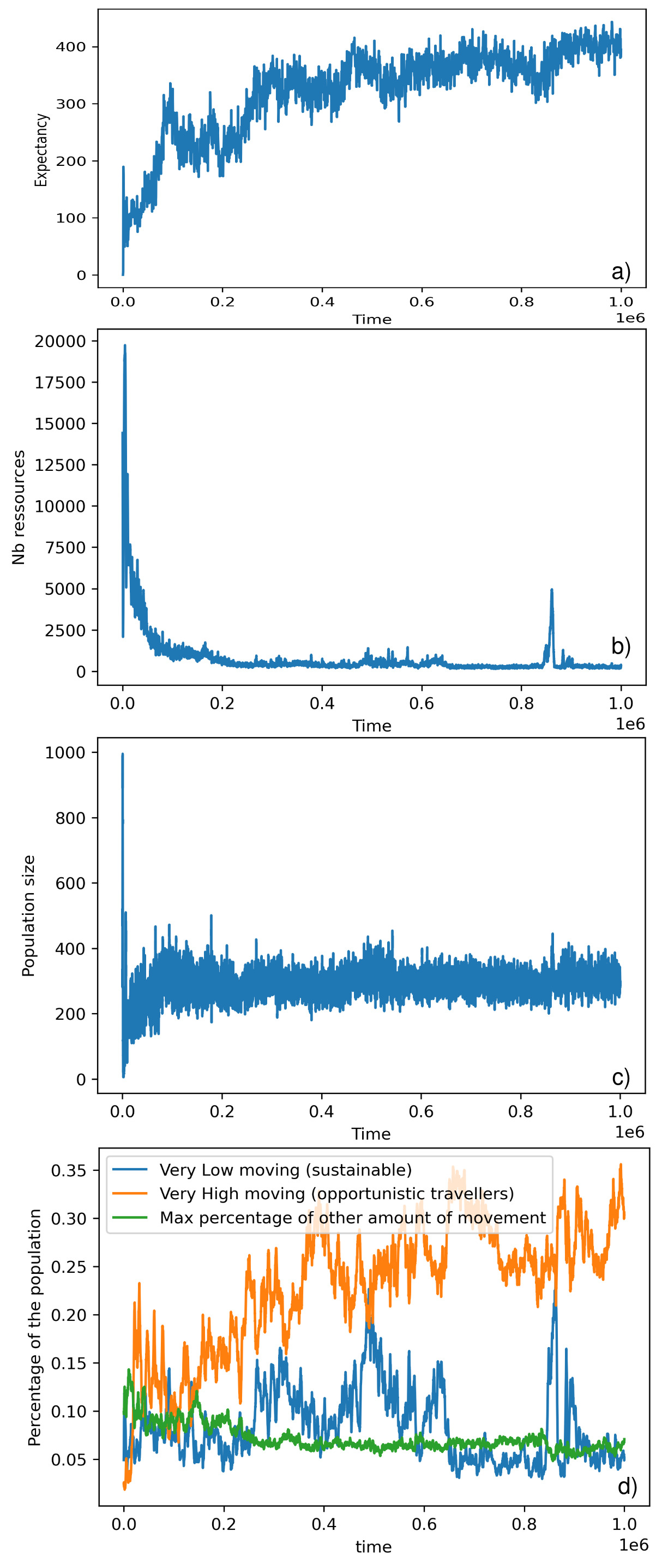}
    \end{center}
\end{minipage}
\caption{Metrics on seed 2: a) life expectancy of agents, b) total number of resources present in the environment, c) size of the population and d) percentage of individuals with different amount of movement}
\label{fig:simulation_metrics_seed2}
\end{figure}

Figure \ref{fig:simulation_metrics_seed2} displays all metrics we discussed in Section \ref{appendix:seed1_metrics} for seed 1, this time measured for seed 2.

\subsubsection{Diversity of eco-evolutionary path}

 \begin{figure}

\begin{minipage}{0.45\textwidth}
    \begin{center}
    \includegraphics[width=\textwidth, height=4cm]{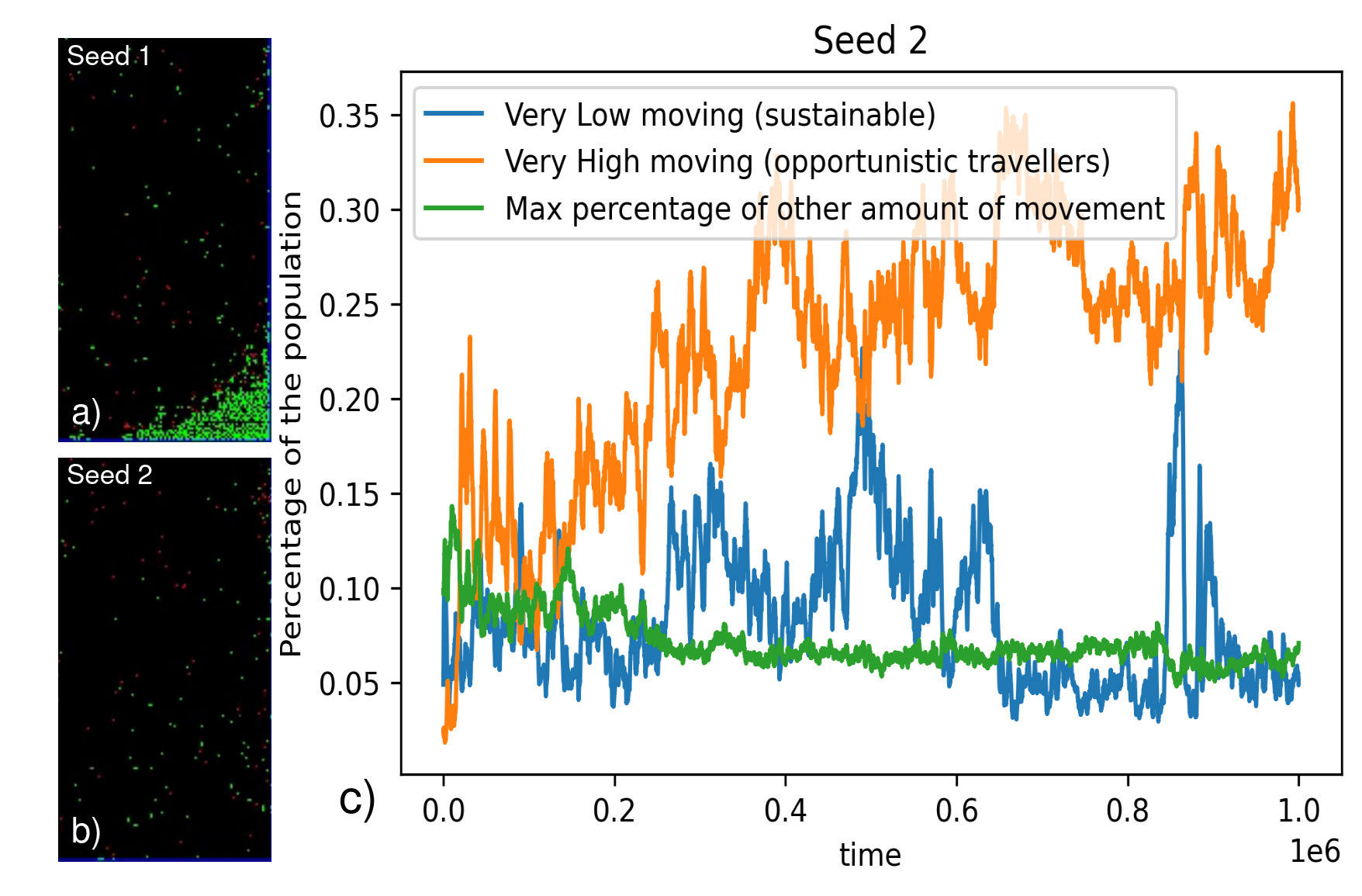} 
    \end{center}
\end{minipage} 

\caption{Left: Diversity of environment between the 2 seeds (at timestep 600 000, zoom on the bottom right corner), here agents are in red; Right: Percentage of the population with different amounts of movement of seed 2 }
\label{fig:evo_path}
\end{figure}

We will now study some differences between seed 1 and 2.

In seed 1 sustainable and  opportunistic travelers coexist during the whole evolution (fig \ref{fig:simulation_metrics}.D), while seed 2 has a majority of opportunistic travelers and some sparse period where low moving agents emerge (fig \ref{fig:evo_path}.c). This may be explained by the differences in the environment led by the agents behavior. In fact seed 1 displays some area where there are big patches of resources (especially in the corner) (fig \ref{fig:evo_path}.a)) and so where sustainable agents can easily take advantage of. On the other end in seed 2 (fig \ref{fig:evo_path}.b), the map is completely depleted of patches of resources which only allows agents to sustain on spontaneous regrowth on random spots of the map, which might explain why there are so much opportunistic travelers and nearly no sustainable behavior. The sustainable behavior pic we can see in seed 2 might be explained by timesteps where some spot of food were left for some time and so where bigger patches of resources emerged which might have favored some switches in behavior. See Videos 1.a and 1.b of the \href{https://sites.google.com/view/non-episodic-neuroevolution-in/}{\underline{companion website}} for a better visualization of the dynamic and behavior of agents. 

The (small) diversity of evolutionary and environment path between the 2 seeds we present are also an interesting feature of such eco-evo simulation.

\subsection{Lab additional results}

In this section, we provide additional results on the lab environment.

\begin{figure}
    \centering
    \includegraphics[width=0.45\textwidth]{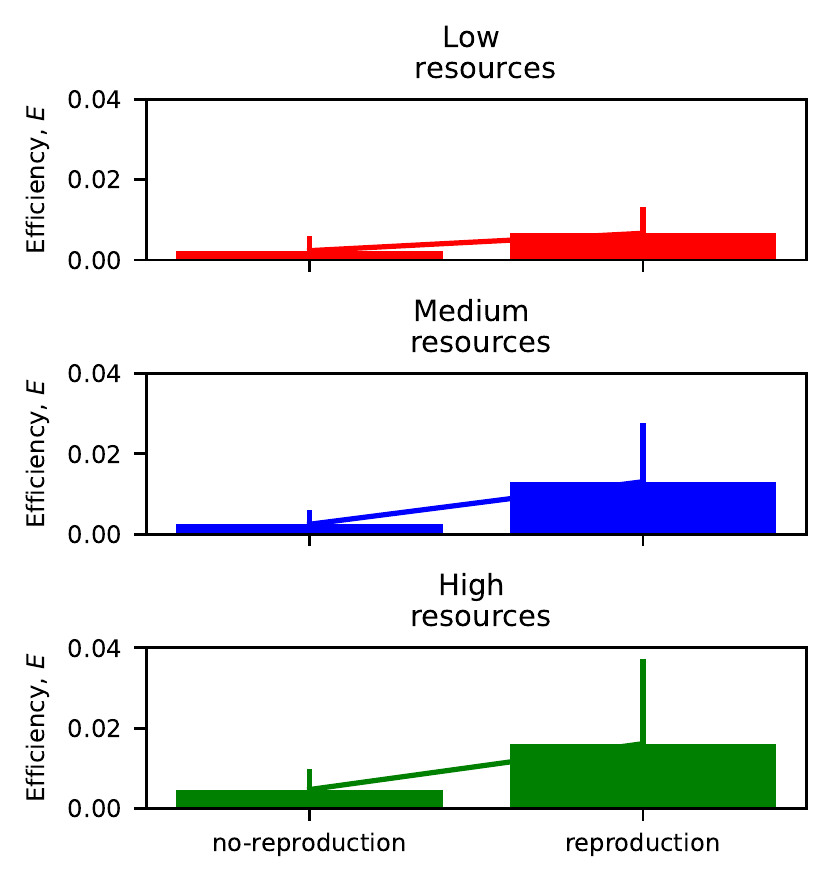}
    \caption{Average efficiency across the population across different density levels with reproduction activated and deactivated. Activating reproduction leads to increased resource consumption.}
    \label{fig:eval_pressure}
\end{figure}

 Fig \ref{fig:eval_pressure} shows the average efficiency averaged on the population on different density of resources (compared to only high resources task in fig \ref{fig:all_results}.D) with reproduction activated and deactived, we observe that on every resource density, activating reproduction leads to increased resource consumption.
 
\end{document}